\documentclass[11pt,a4paper]{article}
\usepackage{times,latexsym}
\usepackage{url}
\usepackage[T1]{fontenc}

\usepackage[acceptedWithA]{tacl2021v1}
\usepackage{tacl2021v1}

\usepackage{xspace,mfirstuc,tabulary}
\newcommand{\dateOfLastUpdate}{Dec. 15, 2021}
\newcommand{\styleFileVersion}{tacl2021v1}

\newcommand{\ex}[1]{{\sf #1}}

\newif\iftaclinstructions
\taclinstructionsfalse 
\iftaclinstructions
\renewcommand{\confidential}{}
\renewcommand{\anonsubtext}{(No author info supplied here, for consistency with
TACL-submission anonymization requirements)}
\newcommand{\instr}
\fi

\iftaclpubformat 
\newcommand{\taclpaper}{final version\xspace}

\newcommand{\Taclpaper}{Final version\xspace}
\newcommand{\Taclpapers}{Final versions\xspace}
\newcommand{\TaclPapers}{Final Versions\xspace}
\else
\newcommand{\taclpaper}{submission\xspace}

\newcommand{\Taclpaper}{Submission\xspace}
\newcommand{\Taclpapers}{{\Taclpaper}s\xspace}
\newcommand{\TaclPapers}{Submissions\xspace}
\fi

\title{Formatting Instructions for TACL \TaclPapers \\
(Base files: \styleFileVersion-template.tex \& \styleFileVersion.sty, dated \dateOfLastUpdate)}

\author{Ting-Yao `Edward' Hsu\textsuperscript{1}~
Yi-Li Hsu\textsuperscript{2}~
Shaurya Rohatgi\textsuperscript{3}~
Chieh-Yang Huang\textsuperscript{4}~
Ho Yin Sam Ng\textsuperscript{1}\\
\textbf{Ryan Rossi\textsuperscript{5}~
Sungchul Kim\textsuperscript{5}~
Tong Yu\textsuperscript{5}~
Lun-Wei Ku\textsuperscript{6}~
Clyde Lee Giles\textsuperscript{1}~Ting-Hao `Kenneth' Huang\textsuperscript{1}}\\
  \textsuperscript{1}Pennsylvania State University~~~
  \textsuperscript{2}National Tsing Hua University, Taiwan~~~
  \textsuperscript{3}AllSci\\
  \textsuperscript{4}MetaMetrics Inc.~~~
  \textsuperscript{5}Adobe Research~~~
  \textsuperscript{6}Institute of Information Science, Academia Sinica\\
\textsuperscript{1}\texttt{\{txh357,sam.ng,clg20,txh710\}@psu.edu}~~~\textsuperscript{3}\texttt{srohatgi@allsci.com}\\
\textsuperscript{4}\texttt{cyhuang@lexile.com}~~~
\textsuperscript{5}\texttt{\{ryrossi,sukim,tyu\}@adobe.com}\\
\textsuperscript{2,6}\texttt{\{yili.hsu,lwku\}@iis.sinica.edu.tw}
}

\usepackage{xspace}
\usepackage{comment}
\usepackage{xcolor}
\usepackage{multirow}
\usepackage{booktabs}
\usepackage{cleveref}
\usepackage{graphicx}
\usepackage[inline]{enumitem}
\usepackage{subcaption}
\usepackage[normalem]{ulem}
\usepackage{siunitx}

\useunder{\uline}{\ul}{}

\title{Do \LMMs Solve Caption Generation for Scientific Figures? Lessons Learned from \textsc{SciCap} Challenge 2023}

\newcommand{\LMMs}{Large Multimodal Models\xspace}
\newcommand{\LMM}{Large Multimodal Model\xspace}

\newcommand{\lmms}{large multimodal models\xspace}

\newcommand{\dataset}{\textsc{SciCap}\xspace}

\newcommand{\eg}{{\it e.g.}\xspace}
\newcommand{\ie}{{\it i.e.}\xspace}

\begin{document}
\maketitle

\begin{abstract}
Since the \dataset dataset's launch in 2021, the research community has made significant progress in generating 
captions for scientific figures in scholarly articles.
In 2023, the first \dataset Challenge took place, inviting global teams to use an expanded \dataset dataset to develop models for captioning diverse figure types across various academic fields.
At the same time, text generation models advanced quickly, with many powerful pre-trained \lmms (LMMs) emerging that showed impressive capabilities in various vision-and-language tasks.
This paper presents an overview of the first \dataset Challenge and details the performance of various models on its data, capturing a snapshot of the field's state.
We found that professional editors overwhelmingly preferred figure captions generated by GPT-4V over those from all other models and even the original captions written by authors.
Following this key finding, we conducted detailed analyses to answer this question: \textbf{Have advanced LMMs solved the task of generating captions for scientific figures?}

\end{abstract}


\section{Introduction and Background\label{sec:introduction}}
Scientists use figures like bar charts, pie charts, or scatter plots to convey key findings in scholarly articles.
However, the texts accompanying these figures, the figure captions, are often overlooked by the authors and do not receive the needed attention when being composed.
Even though many studies have shown the role of captions in enhancing readers' comprehension and recall of the messages conveyed by figures~\cite{nugent1983deaf,large1995multimedia,bransford1979human,hegarty1993constructing}, poorly-written captions are, unfortunately, common~\cite{huang2023summaries}.


\begin{figure}
    \centering
    \includegraphics[width=0.99\linewidth]{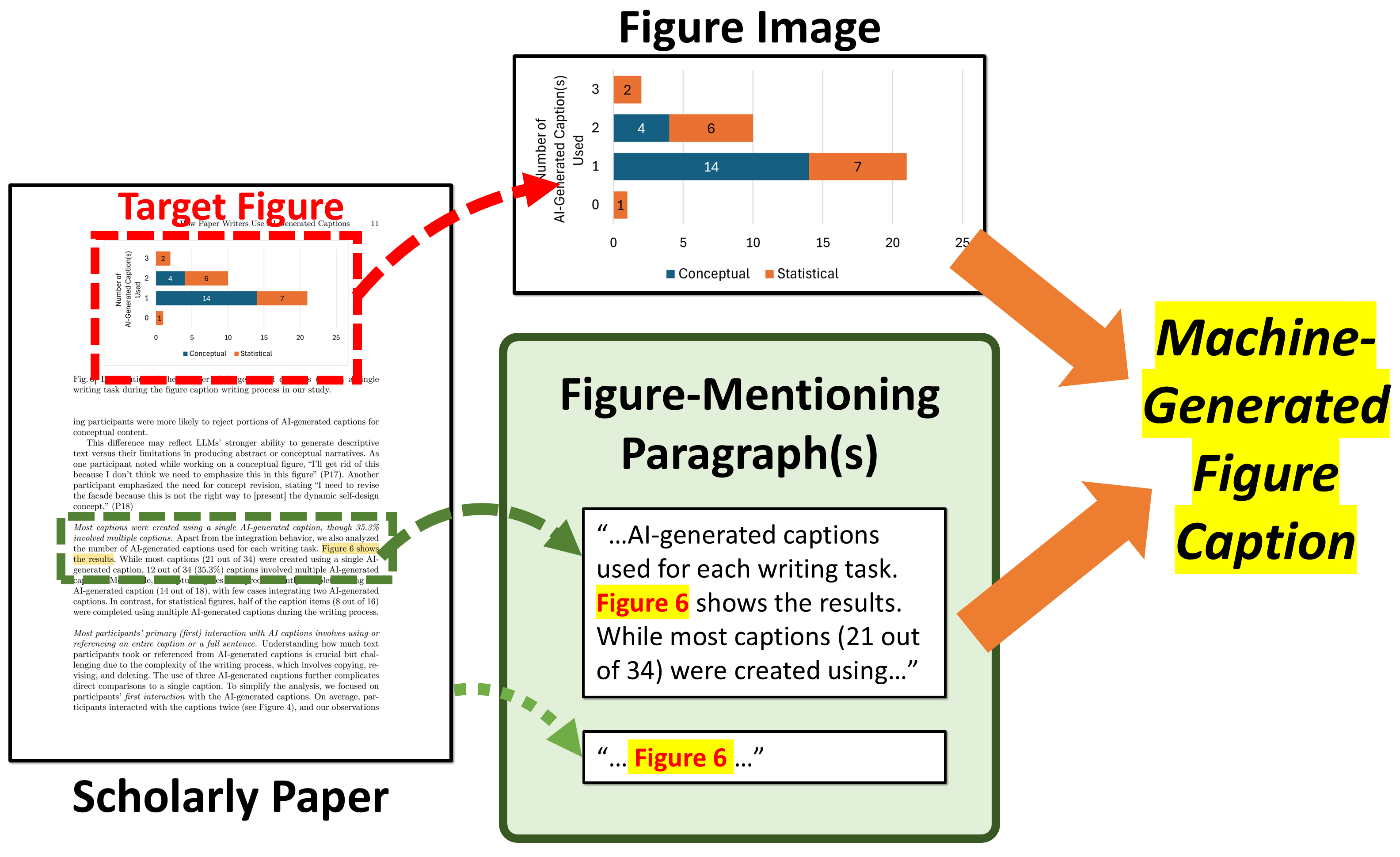}
    \vspace{-1pc}
    \caption{In \dataset Challenge, models generate captions based on the figure and the figure-mentioning paragraph. [Paper source:~\citet{ngunderstanding}]}
    \vspace{-1pc}
    \label{fig:overview}
\end{figure}

In response, \citet{hsu2021scicap} launched \dataset, a large-scale collection of
133,543 single-panel line charts and their captions extracted from arXiv papers, aiming to fuel the creation of new models that generate high-quality captions for scientific figures.
Over the past three years, \dataset has advanced researchers' understanding of scientific figure captions and driven the progress of technologies generating them:
\dataset confirmed that many low-quality captions exist in scholarly articles, as over half of the figure captions in arXiv {\tt cs.CL} papers were rated ``unhelpful'' by NLP Ph.D. students~\cite{huang2023summaries};
\dataset revealed that producing figure captions in scholarly articles is a generative task heavily reliant on the texts within the articles~\cite{yang2023scicap+,li2023scigraphqa,horawalavithana2023scitune}--
this task relies so much on the paper content that it can be more effectively tackled through text summarization, summarizing all paragraphs mentioning the figures (\eg, ``Figure~3 shows...''), rather than as a vision-to-language task~\cite{huang2023summaries};
\dataset elevated the quality of generated captions to a level where, in instances where author-created captions were poorly crafted, readers found the generated captions more helpful~\cite{huang2023summaries}, thereby offering practical assistance in caption writing~\cite{scicapenter2024}.
In 2023, the first \dataset Challenge took place.
With an expanded \dataset of 476,389 single-panel figures from 8 domains and 5 figure types, the challenge invited teams worldwide to develop caption-generation models for diverse figures.

Meanwhile, in the past two years, the landscape of text generation has rapidly evolved:
ChatGPT (GPT-3.5) was initially released to the public in November 2022;
its more powerful successor, GPT-4, was released in March 2023; 
and GPT-4V, the successor to GPT-4, which can take images and texts as input, became publicly available in September 2023.
Numerous open-source pre-trained Large Language Models (LLMs) and \LMMs (LMMs), such as OPT~\cite{zhang2022opt}, LLaMA-2~\cite{touvron2023llama}, Mistral~\cite{jiang2023mistral}, Gemma~\cite{team2024gemma}, LLaVA~\cite{liu2023llava}, BLIP-2~\cite{li2023blip2}, mPLUG-Owl2~\cite{ye2023mplug} and MiniGPT-4~\cite{zhu2023minigpt}, have been made available.
All these models have shown impressive progress in understanding charts and tables within documents~\cite{yue2023mmmu}.
Now is a good time 
to take a step back and critically assess the collective advancements made in generating high-quality, useful captions for scientific figures.
Specifically, we aim to answer this question: 
\textbf{Have the impressive \lmms solved the challenge of generating good captions for scientific figures?} 

This paper first overviews the 2023's \dataset Challenge, including its data, procedure, winning teams and their models, which represents the status quo of the scientific figure caption generation task (\Cref{sec:challenge}).
Next, we ran GPT-4V---unavailable at the time of the challenge---and other open LMMs, such as UniChart~\cite{masry2023unichart}, on \dataset Challenge's data and, through automatic and human evaluations, compared their performances to all models that participated in the challenge (\Cref{sec:model-breakdown}).
The automatic evaluation revealed that the linguistic features of captions differ by domain and figure type, resulting in a model's scores varying across these categories.
More interestingly, in the human evaluation, three professional editors with expertise in technical academic writing \textbf{unanimously preferred captions generated by GPT-4V over those from all other models, including the original captions written by the authors}.
Driven by this key finding, we investigated why editors strongly prefer GPT-4V (\Cref{sec:comment-analysis}), comparing their views with those of Ph.D. and undergrad students (\Cref{sec:student-ratings}), to directly address whether LMMs have solved scientific figure caption generation.
We concluded by identifying unresolved challenges and suggesting future research directions (\Cref{sec:discussion}).

\section{Related Work\label{sec:backgrounds}}
In addition to the works directly related to \dataset (Section~\ref{sec:introduction}), the complex nature of figure narrating---which demands an understanding of both vision and text, along with intricate domain-specific contexts---has made it an intellectually intriguing challenge, inspiring various parallel projects focused on text generation for figures.
FigJAM leveraged metadata and a combined static and dynamic dictionary to generate better caption units~\cite{qian2021generating}. 
FigCAP developed innovative attention mechanisms and employed reinforcement learning for sequence-level training to enhance caption generation performance~\cite{chen2020figure}.
Large-scale chart-to-text benchmark datasets have also been created to facilitate data-to-text techniques for chart summarization~\cite{kantharaj2022chart,obeid2020chart}.
\citet{Tarsi_2024_WACV} presented two datasets for scientific multimodal learning: 
SciOL, a large corpus covering various sciences, and 
MuLMS-Img, a high-quality annotated materials science dataset. 
\citet{xia2024chartx} introduced ChartX, a comprehensive evaluation set with 18 chart types, 7 tasks, 22 topics, and developed ChartVLM, a model tailored for multi-modal tasks requiring interpretable patterns, like chart reasoning. 
Unichart developed specialized pretraining tasks for charts, focusing on visual element extraction and reasoning skills~\cite{masry2023unichart}. 
ChartSumm introduced a comprehensive benchmark dataset with 84,363 charts, metadata, and descriptions across various topics and types for creating concise and detailed summaries~\cite{rahman2023chartsumm}. 
While inspiring, most efforts did not aim to produce real-world captions from scholarly articles.
To show this, we included UniChart's output in our study.

\section{The First \dataset Challenge (2023)\label{sec:challenge}}

The first \dataset Challenge was held at the 5th Workshop on Closing the Loop Between Vision and Language during ICCV 2023.
The organizers expanded the \dataset dataset to produce a challenge-specific version
which was divided into test, validation, and training sets, containing 476,389 single-panel arXiv figures. 
Teams had three months to develop solutions and submit model outputs for all test set figures.
This section details the challenge.

\subsection{Challenge Dataset\label{sec:data-prepare}}

The \dataset challenge dataset comprises 476,389 single-panel figures from arXiv papers\footnote{The organizing team also supplied 26,868 figures from ACL papers as extra training data, which were not included in the data mentioned in this section.}
in 8 domains: 
{\em (i)} Computer Science (\texttt{cs}), 
{\em (ii)} Economics (\texttt{econ}), 
{\em (iii)} Electrical Engineering and Systems Science (\texttt{eess}), 
{\em (iv)} Mathematics (\texttt{math}), 
{\em (v)} Physics (\texttt{physics}), 
{\em (vi)} Quantitative Biology (\texttt{q-bio}), 
{\em (vii)} Quantitative Finance (\texttt{q-fin}), 
and 
{\em (viii)} Statistics (\texttt{stat}). 
It covers 5 figure types, as denoted by the figure type classifier, FigureSeer~\cite{siegel2016figureseer}: 
{\em (i)} Node Diagram, 
{\em (ii)} Equation, 
{\em (iii)} Graph Plot, 
{\em (iv)} Scatterplot, and 
{\em (v)} Bar Chart. 
This dataset, with 476,389 figures, is more than 3 times the size of the original \dataset dataset, which contains 133,543 figures.
The figures were randomly sampled from arXiv papers published between 2010 and 2020, with a maximum of four figures selected per paper. 
While the organizers had access to the full collection of over 440,000 arXiv papers from this period, they intentionally limited the dataset size to encourage participation for teams with limited computational resources.
\Cref{app:data-prepare} details the dataset preparation process.

Concerns about the quality of arXiv papers are understandable. 
However, \citet{huang2023summaries} manually verified 399 figures from the original \dataset test set and found that 81.2\% (324/399) originated from papers published at academic conferences, with 51.9\% (207/399) appearing in the ACL Anthology, IEEE, or ACM.\footnote{These results were reported in the paper's Appendix B.} 
These findings suggest that the dataset is both representative and of reasonable quality.




\paragraph{Data Split.}
The dataset was divided as follows: 
training set (70\%, 333,472 figures), 
validation set (10\%, 47,639 figures), 
public test set (10\%, 47,639 figures), and 
hidden test set (10\%, 47,639 figures). 
Teams can submit results for the public test set throughout the three-month submission period to refine their systems.
Access to the hidden test set was granted in the final two weeks, during which teams could submit their hidden test set results. 
The winning teams were determined based on performance on the hidden test set.

\paragraph{Selecting the ``Quality Subset'' from the Hidden Test Set.}
To address issues with unreliable data skewing reference-based evaluations, the \dataset Challenge organizers curated a ``Quality Subset'' of the hidden test set, composed only of high-quality captions. 
Starting with a random sample of 1,000 figures from the hidden test set, they applied \citet{hsu2023gpt}'s approach to have GPT-4 assign each caption a quality score between 1 and 6.
Captions scoring below 5 were discarded, resulting in 769 captions. 
The organizers then manually evaluated these captions, ultimately selecting 460 captions for the ``Quality Subset.''
This subset was used to determine the Quality Winner.

\paragraph{Data Provided.}
For each figure in the training, validation, and public test set, the following data items were provided:
{\em (i)} figure image,
{\em (ii)} figure caption, {\em (iii)} figure type , {\em (iv)} OCR (textual information extracted from the image) and {\em (v)} the paragraphs that mentioned the figure (\eg, ``As shown in Figure 5, ...''.)
All of these were extracted from the original arXiv paper.
For each figure in the hidden test set, only the image and the paragraphs, but not the caption,
were provided.

\paragraph{Full-Scale \dataset Data Release.}
Following the \dataset Challenge 2023 and 2024~\cite{kim2025multi},
and in response to growing interest in large-scale figure caption data, the organizing team decided to release the full \dataset dataset, including all figure captions extracted from arXiv papers published between 2010 and 2020.\footnote{\dataset Full Data Release: \url{https://huggingface.co/datasets/CrowdAILab/Scicap-All}}
Future \dataset Challenges will use newly published arXiv papers as test and hidden test data.

\subsection{Challenge Procedure\label{sec:challenge-procedure}}

The SciCap Challenge was hosted on EvalAI~\cite{yadav2019evalai}.
It had two phases: the \textbf{Test Phase} and the \textbf{Challenge Phase}. 
During the Test Phase---which began on May 29, 2023, and lasted approximately 2.5 months---each team used the provided training set, validation set, and public test set to build and test their models. 
Each team can upload their predictions for the public test set to Eval.AI, which automatically calculates the scores using a series of evaluation metrics. 
Then, in the Challenge Phase, which took place in the final 2 weeks before the submission deadline (August 15-31, 2023), the hidden test set was released. 
Teams can then submit their results for the hidden test set. 
Each team can submit up to five times per day during the challenge phase.
To qualify for the winners, the team must submit their prediction outputs for the hidden test set and provide a two to four-page technical report. 

\paragraph{Rules for Using External Data and Models.}
Teams were allowed to use LLMs like GPT-4 or LLaMA and any external data for their captioning systems. 
However, using the original captions from the hidden test set was prohibited to maintain fairness.
Although we did not provide gold captions in the hidden test set, arXiv papers are openly accessible, making it impossible to fully prevent potential data leakage. 
Teams caught using this gold data will be disqualified. 
Each team must clearly describe their methodology in the technical report to enable replication.

\paragraph{Evaluation and Winning Criteria.}
The challenge selected two winners: the \textbf{Leaderboard Winner} and the \textbf{Quality Winner}.\footnote{The challenge was set to award two distinct winning teams. If one team tops both evaluations, the runner-up in the Quality Winner track becomes the winner for that category.}
The challenge's leaderboard reported seven automatic scores that were commonly used in prior works~\cite{huang2023summaries,hsu2021scicap}, using author-written captions as references: 
BLEU-4, ROUGE-1, ROUGE-2, ROUGE-L, and the normalized versions of ROUGE scores.
Note that the ROUGE's F1 scores were used.
The \textbf{ROUGE-2-Normalized} score was chosen as the deciding metric for determining the winners, as it addresses the issue of standard ROUGE F1 favoring longer texts~\cite{huang2023summaries,sun-etal-2019-compare}.
The Leaderboard Winner achieved the highest ROUGE-2-Normalized score on the entire hidden test set, and the Quality Winner achieved the highest ROUGE-2-Normalized score on the specially curated Quality Subset of the hidden test set (Section~\ref{sec:data-prepare}).
The Quality Winner track was inspired by \citet{huang2023summaries}, highlighting that a model performing well on a full test set may struggle with higher-quality captions.

\begin{table}[t]
\small
\centering
\resizebox{\linewidth}{!}{
\begin{tabular}{@{}l@{}c@{\kern4pt}c@{\kern4pt}c@{\kern4pt}c@{\kern4pt}c@{\kern4pt}c@{\kern4pt}c@{}}
\toprule
\multirow{2}{*}{\textbf{Team}} & \multirow{2}{*}{\textbf{BLEU-4}} & \multicolumn{2}{@{}c@{}}{\textbf{ROUGE-1}} & \multicolumn{2}{@{}c@{}}{\textbf{ROUGE-2}} & \multicolumn{2}{@{}c@{}}{\textbf{ROUGE-L}} \\ \cmidrule{3-8}
         &    & \textbf{Score} & \textbf{Norm} & \textbf{Score} & \textbf{Norm} & \textbf{Score} & \textbf{Norm}\\ \midrule
         
\multicolumn{8}{@{}c@{}}{\textbf{Hidden Set}} \\ \midrule
NJUST                  & .082     & \textbf{.406}  & \textbf{2.331}     & \textbf{.250} & \textbf{4.489}            & \textbf{.378} & \textbf{2.617}     \\
USTC                   & .043            & .297 & 1.613      & .145 & 2.418            & .256 & 1.700    \\
\midrule
Pegasus  & \textbf{.099}     & \underline{.363} & \underline{1.880}      & \underline{.220} & \underline{3.462}            & \underline{.325} & \underline{2.086}
\\
Pegasus\textsuperscript{*}  & \underline{.088}     & .340 & 1.832      & .200 & 3.313            & .304 & 2.001 \\ \midrule

\multicolumn{8}{@{}c@{}}{\textbf{Quality Subset}} \\ \midrule 
NJUST  & .070 & \underline{.372}  & \textbf{2.043} & \underline{.226} & \textbf{3.828}   & \textbf{.338} & \textbf{2.260}     \\
USTC & .049 & .307 & 1.612 & .157 & 2.503  & .260 & 1.680    \\
\midrule
Pegasus  & \textbf{.118}     & \textbf{.384} & \underline{1.887}      & \textbf{.240} & \underline{3.507}            & \textbf{.338} & \underline{2.090}
\\
Pegasus\textsuperscript{*}  & \underline{.096}     & .358 & 1.825      & .214 & 3.293            & .312 & 1.981
\\
\bottomrule
\end{tabular}
}
\vspace{-.5pc}
\caption{Leaderboard scores of the two winning teams and the two baselines. The \textbf{best} and \underline{second-best} results are highlighted. Note that Pegasus is finetuned on the Scicap Challenge dataset and Pegasus\textsuperscript{*} is finetuned on graphplots only~\cite{huang2023summaries}.}
\vspace{-.8pc}
\label{leaderboard}
\end{table}

\subsection{Challenge Results and Winning Teams}
Team NJUST-KMG~\cite{winningteam} scored highest on both public and hidden test sets, with Team USTC-IAT-United~\cite{qualityteam} as the runner-up in the Quality Winner track. 
According to the rules, NJUST-KMG was named the Leaderboard Winner, and USTC-IAT-United was selected as the Quality Winner. 
Table~\ref{leaderboard} shows the evaluation scores of the winning teams on the entire Hidden Test Set and the Quality Subset, including scores from two baseline models the \dataset Challenge reported in the leaderboard.

\paragraph{Leaderboard Winner: NJUST-KMG~\cite{winningteam}.}
The team employed the PP-OCRv3 model from PaddleOCR~\cite{du2020pp}
to extract precise information from images.
They used LLaMA-2-7B to refine the paragraphs with summarization.
This refined dataset was then used to train a PegasusX model~\cite{phang2022investigating} to generate candidate summaries.
Another BRIO model~\cite{liu-etal-2022-brio} was trained with these summaries, using contrastive learning to optimize ROUGE-2-normalized performance.

\paragraph{Quality Winner: USTC-IAT-United~\cite{qualityteam}.}
The team first enriched the dataset by concatenating BLIP-v2~\cite{li2023blip2} captions, paragraphs, and mentions. 
Then a Pegasus model~\cite{zhang2020pegasus} was finetuned on this enriched dataset.
Although the team tried different summarization models, the fine-tuned Pegasus was shown to be the most effective one.

\section{In-Depth Evaluation of Models' Performances\label{sec:model-breakdown}}


\begin{table*}[ht!]
\centering
\mysmall
\begin{tabular}{@{}lcS[table-format = 2.1, detect-all]ccccccc@{}}
\toprule
\multirow{2}{*}{\textbf{Model}} & \multirow{2}{*}{\textbf{Domain}} & \multicolumn{1}{c}{\multirow{2}{*}{\shortstack{\textbf{Token}\\\textbf{Length}}}}
 & \multirow{2}{*}{\textbf{BLEU-4}} & \multicolumn{2}{c}{\textbf{ROUGE-1}} & \multicolumn{2}{c}{\textbf{ROUGE-2}} & \multicolumn{2}{c}{\textbf{ROUGE-L}} \\ \cmidrule(lr){5-6} \cmidrule(lr){7-8} \cmidrule(lr){9-10} 
 &  &  &  & \textbf{Score} & \textbf{Norm} & \textbf{Score} & \textbf{Norm} & \textbf{Score} & \textbf{Norm} \\ \midrule
\multirow{8}{*}{\textbf{NJUST}} & \textbf{physics} & 14.4 & {\ul .062} & .323 & 1.856 & {\ul .195} & 3.494 & {\ul .287} & 1.986 \\
 & \textbf{cs} & 8.3 & {\ul .078} & {\ul .373} & 2.950 & {\ul .230} & 6.317 & {\ul .349} & 3.143 \\
 & \textbf{math} & 9.5 & {\ul .070} & {\ul .339} & 2.447 & {\ul .205} & 4.975 & {\ul .317} & 2.638 \\
 & \textbf{q-bio} & 11.4 & {\ul .062} & .307 & 1.990 & {\ul .190} & 3.991 & {\ul .279} & 2.129 \\
 & \textbf{stat} & 9.9 & {\ul .068} & .334 & 2.348 & {\ul .201} & 4.691 & {\ul .310} & 2.520 \\
 & \textbf{q-fin} & 10.3 & {\ul .053} & .319 & 2.191 & {\ul .193} & 4.385 & {\ul .293} & 2.347 \\
 & \textbf{eess} & 9.3 & {\ul .100} & {\ul .414} & 3.031 & {\ul .260} & 6.431 & {\ul .386} & 3.260 \\
 & \textbf{econ} & 10.0 & {\ul .067} & {\ul .361} & 2.534 & {\ul .214} & 4.985 & {\ul .331} & 2.690 \\ \hline
\multirow{8}{*}{\textbf{USTC}} & \textbf{physics} & 21.6 & .050 & .319 & 1.584 & .156 & 2.316 & .268 & 1.665 \\
 & \textbf{cs} & 10.9 & .048 & .314 & 2.095 & .155 & 3.378 & .284 & 2.212 \\
 & \textbf{math} & 12.8 & .043 & .308 & 1.871 & .150 & 2.892 & .277 & 2.008 \\
 & \textbf{q-bio} & 15.9 & .042 & .291 & 1.603 & .140 & 2.373 & .250 & 1.674 \\
 & \textbf{stat} & 13.1 & .048 & .306 & 1.838 & .149 & 2.829 & .271 & 1.940 \\
 & \textbf{q-fin} & 14.1 & .044 & .295 & 1.712 & .150 & 2.732 & .261 & 1.818 \\
 & \textbf{eess} & 11.4 & .061 & .347 & 2.245 & .178 & 3.734 & .315 & 2.394 \\
 & \textbf{econ} & 13.2 & .048 & .321 & 1.926 & .156 & 2.953 & .291 & 2.078 \\ \hline
\multirow{8}{*}{\textbf{Pegasus}} & \textbf{physics} & 24.4 & \textbf{.109} & \textbf{.384} & 1.859 & \textbf{.232} & 3.312 & \textbf{.334} & 2.045 \\
 & \textbf{cs} & 12.4 & \textbf{.118} & \textbf{.394} & 2.448 & \textbf{.244} & 4.843 & \textbf{.367} & 2.699 \\
 & \textbf{math} & 14.3 & \textbf{.113} & \textbf{.391} & 2.259 & \textbf{.243} & 4.389 & \textbf{.364} & 2.525 \\
 & \textbf{q-bio} & 19.8 & \textbf{.106} & \textbf{.367} & 1.870 & \textbf{.229} & 3.525 & \textbf{.329} & 2.086 \\
 & \textbf{stat} & 16.3 & \textbf{.122} & \textbf{.384} & 2.092 & \textbf{.242} & 4.074 & \textbf{.352} & 2.338 \\
 & \textbf{q-fin} & 16.2 & \textbf{.102} & \textbf{.361} & 1.970 & \textbf{.228} & 3.837 & \textbf{.329} & 2.188 \\
 & \textbf{eess} & 12.8 & \textbf{.133} & \textbf{.428} & 2.605 & \textbf{.269} & 5.186 & \textbf{.397} & 2.870 \\
 & \textbf{econ} & 17.8 & \textbf{.152} & \textbf{.439} & 2.314 & \textbf{.292} & 4.695 & \textbf{.409} & 2.651 \\ \hline
\multirow{8}{*}{\shortstack[l]{\textbf{Unichart}\\\textbf{(Finetuned)}}} & \textbf{physics} & 20.3 & .005 & .174 & 0.879 & .049 & 0.738 & .148 & 0.932 \\
 & \textbf{cs} & 8.1 & .007 & .184 & 1.481 & .065 & 1.830 & .168 & 1.539 \\
 & \textbf{math} & 13.1 & .004 & .163 & 0.981 & .045 & 0.851 & .149 & 1.068 \\
 & \textbf{q-bio} & 13.2 & .008 & .185 & 1.107 & .058 & 1.092 & .159 & 1.137 \\
 & \textbf{stat} & 11.1 & .005 & .176 & 1.159 & .056 & 1.194 & .156 & 1.207 \\
 & \textbf{q-fin} & 13.6 & .006 & .177 & 1.041 & .055 & 1.012 & .157 & 1.106 \\
 & \textbf{eess} & 9.3 & .013 & .219 & 1.609 & .081 & 2.007 & .200 & 1.697 \\
 & \textbf{econ} & 12.5 & .003 & .154 & 0.950 & .050 & 0.990 & .141 & 1.033 \\ \hline
\multirow{8}{*}{\shortstack[l]{\textbf{GPT-4V}\\\textbf{(Image)}}} & \textbf{physics} & 20.7 & .004 & .203 & 1.019 & .049 & 0.742 & .157 & 0.984 \\
 & \textbf{cs} & 17.5 & .005 & .219 & 1.159 & .059 & 0.960 & .179 & 1.165 \\
 & \textbf{math} & 17.6 & .002 & .168 & 0.889 & .034 & 0.548 & .139 & 0.906 \\
 & \textbf{q-bio} & 18.2 & .005 & .213 & 1.114 & .059 & 0.930 & .168 & 1.081 \\
 & \textbf{stat} & 17.5 & .003 & .214 & 1.135 & .053 & 0.864 & .170 & 1.106 \\
 & \textbf{q-fin} & 18.0 & .003 & .191 & 0.999 & .046 & 0.737 & .154 & 0.996 \\
 & \textbf{eess} & 18.1 & .005 & .234 & 1.226 & .062 & 0.991 & .191 & 1.233 \\
 & \textbf{econ} & 18.8 & .004 & .212 & 1.100 & .067 & 1.055 & .183 & 1.171 \\ \hline
\multirow{8}{*}{\shortstack[l]{\textbf{GPT-4V}\\\textbf{(Image+Paragraph)}}} & \textbf{physics} & 41.3 & .025 & {\ul .352} & 1.529 & .132 & 1.582 & .258 & 1.480 \\
 & \textbf{cs} & 28.0 & .017 & .324 & 1.537 & .124 & 1.704 & .257 & 1.559 \\
 & \textbf{math} & 30.9 & .020 & .313 & 1.458 & .119 & 1.582 & .249 & 1.489 \\
 & \textbf{q-bio} & 35.0 & .020 & {\ul .333} & 1.509 & .122 & 1.564 & .245 & 1.439 \\
 & \textbf{stat} & 31.4 & .018 & {\ul .336} & 1.559 & .125 & 1.662 & .256 & 1.527 \\
 & \textbf{q-fin} & 31.0 & .024 & {\ul .328} & 1.527 & .127 & 1.698 & .257 & 1.535 \\
 & \textbf{eess} & 29.6 & .022 & .335 & 1.573 & .130 & 1.766 & .268 & 1.613 \\
 & \textbf{econ} & 31.3 & .022 & .330 & 1.530 & .130 & 1.733 & .266 & 1.588 \\ \bottomrule
\end{tabular}
\vspace{-.5pc}
\caption{Results breakdown by domain for each model, highlighting performance metrics such as BLEU-4, ROUGE scores, and token length. We show the models included in the human evaluation process. The \textbf{highest} and \underline{second highest} values across models are highlighted.}
\vspace{-.5pc}
\label{tab:results_by_domain}
\end{table*}

\begin{table*}[ht!]
\centering \mysmall
\begin{tabular}{@{}lcS[table-format = 2.1, detect-all]ccccccc@{}}
\toprule
\multirow{2}{*}{\textbf{Model}} & \multirow{2}{*}{\textbf{Domain}} & \multicolumn{1}{c}{\multirow{2}{*}{\shortstack{\textbf{Token}\\\textbf{Length}}}}
 & \multirow{2}{*}{\textbf{BLEU-4}} & \multicolumn{2}{c}{\textbf{ROUGE-1}} & \multicolumn{2}{c}{\textbf{ROUGE-2}} & \multicolumn{2}{c}{\textbf{ROUGE-L}} \\ \cmidrule(lr){5-6} \cmidrule(lr){7-8} \cmidrule(lr){9-10} 
 &  &  &  & \textbf{Score} & \textbf{Norm} & \textbf{Score} & \textbf{Norm} & \textbf{Score} & \textbf{Norm} \\ \midrule
\multirow{5}{*}{\textbf{NJUST}} & \textbf{Node Diagram} & 6.5 & {\ul .079} & {\ul .359} & 3.452 & {\ul .231} & 8.369 & {\ul .342} & 3.657 \\
 & \textbf{Equation} & 8.6 & {\ul .068} & {\ul .341} & 2.650 & {\ul .212} & 5.690 & {\ul .318} & 2.826 \\
 & \textbf{Graph Plot} & 13.0 & {\ul .068} & .342 & 2.066 & {\ul .206} & 3.950 & {\ul .309} & 2.223 \\
 & \textbf{Scatterplot} & 13.5 & {\ul .057} & .321 & 1.901 & {\ul .190} & 3.544 & {\ul .284} & 2.010 \\
 & \textbf{Bar Chart} & 10.0 & {\ul .080} & {\ul .380} & 2.666 & {\ul .230} & 5.355 & {\ul .350} & 2.850 \\ \hline
\multirow{5}{*}{\textbf{USTC}} & \textbf{Node Diagram} & 9.2 & .050 & .304 & 2.252 & .154 & 3.869 & .281 & 2.391 \\
 & \textbf{Equation} & 13.4 & .043 & .295 & 1.750 & .141 & 2.642 & .263 & 1.868 \\
 & \textbf{Graph Plot} & 18.5 & .051 & .321 & 1.669 & .158 & 2.499 & .275 & 1.769 \\
 & \textbf{Scatterplot} & 19.9 & .042 & .309 & 1.568 & .144 & 2.215 & .257 & 1.630 \\
 & \textbf{Bar Chart} & 13.2 & .053 & .338 & 2.025 & .165 & 3.131 & .298 & 2.130 \\ \hline
\multirow{5}{*}{\textbf{Pegasus}} & \textbf{Node Diagram} & 11.6 & \textbf{.140} & \textbf{.406} & 2.611 & \textbf{.269} & 5.595 & \textbf{.386} & 2.919 \\
 & \textbf{Equation} & 15.7 & \textbf{.113} & \textbf{.379} & 2.095 & \textbf{.233} & 3.990 & \textbf{.347} & 2.334 \\
 & \textbf{Graph Plot} & 20.7 & \textbf{.110} & \textbf{.387} & 1.946 & \textbf{.234} & 3.531 & \textbf{.344} & 2.156 \\
 & \textbf{Scatterplot} & 22.8 & \textbf{.098} & \textbf{.372} & 1.819 & \textbf{.220} & 3.189 & \textbf{.322} & 1.989 \\
 & \textbf{Bar Chart} & 14.6 & \textbf{.117} & \textbf{.402} & 2.289 & \textbf{.241} & 4.282 & \textbf{.366} & 2.518 \\ \hline
\multirow{5}{*}{\shortstack[l]{\textbf{Unichart}\\\textbf{(Finetuned)}}} & \textbf{Node Diagram} & 5.7 & .002 & .112 & 1.202 & .027 & 1.134 & .107 & 1.256 \\
 & \textbf{Equation} & 9.1 & .001 & .097 & 0.723 & .019 & 0.472 & .090 & 0.768 \\
 & \textbf{Graph Plot} & 17.1 & .007 & .198 & 1.061 & .064 & 1.056 & .172 & 1.131 \\
 & \textbf{Scatterplot} & 21.1 & .004 & .180 & 0.899 & .050 & 0.751 & .154 & 0.962 \\
 & \textbf{Bar Chart} & 9.9 & .008 & .222 & 1.567 & .084 & 1.975 & .201 & 1.645 \\ \hline
\multirow{5}{*}{\shortstack[l]{\textbf{GPT-4V}\\\textbf{(Image)}}} & \textbf{Node Diagram} & 16.7 & .001 & .170 & 0.918 & .034 & 0.562 & .141 & 0.931 \\
 & \textbf{Equation} & 18.0 & .001 & .149 & 0.781 & .027 & 0.423 & .122 & 0.791 \\
 & \textbf{Graph Plot} & 19.9 & .005 & .222 & 1.125 & .059 & 0.904 & .174 & 1.102 \\
 & \textbf{Scatterplot} & 20.8 & .004 & .202 & 1.012 & .049 & 0.735 & .155 & 0.970 \\
 & \textbf{Bar Chart} & 16.6 & .005 & .238 & 1.286 & .069 & 1.148 & .197 & 1.301 \\ \hline
\multirow{5}{*}{\shortstack[l]{\textbf{GPT-4V}\\\textbf{(Image+Paragraph)}}} & \textbf{Node Diagram} & 29.0 & .014 & .297 & 1.402 & .112 & 1.525 & .239 & 1.443 \\
 & \textbf{Equation} & 35.9 & .010 & .255 & 1.146 & .087 & 1.099 & .200 & 1.174 \\
 & \textbf{Graph Plot} & 36.5 & .025 & {\ul .353} & 1.583 & .135 & 1.697 & .266 & 1.552 \\
 & \textbf{Scatterplot} & 40.2 & .024 & {\ul .354} & 1.547 & .132 & 1.604 & .259 & 1.490 \\
 & \textbf{Bar Chart} & 27.7 & .019 & .351 & 1.667 & .133 & 1.830 & .277 & 1.678 \\ \bottomrule
\end{tabular}
\vspace{-.5pc}
\caption{Results breakdown by the figure type for each model, highlighting performance metrics such as BLEU-4, ROUGE scores, and token length. We show the models included in the human evaluation. The \textbf{highest} and \underline{second highest} values across models are highlighted.}
\vspace{-.5pc}
\label{tab:results_by_type}
\end{table*}

\begin{figure*}[t]
    \centering
    \includegraphics[width=\textwidth]{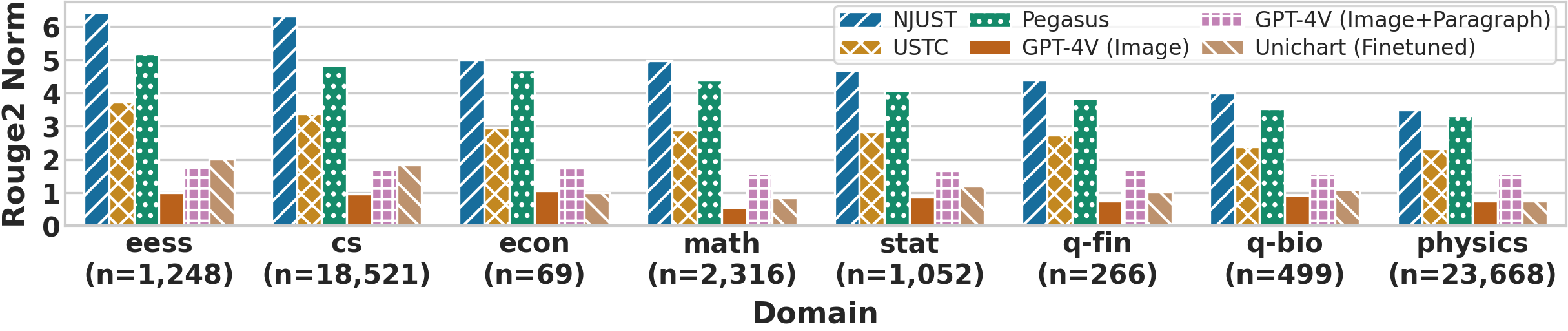}
    \caption{ROUGE-2 normalized scores of each model across eight arXiv domains, highlighting similar trends and demonstrating the generalizability of the caption generation approaches.}
    \label{fig:domains-rouge2_scores}
\end{figure*}

\begin{figure*}[t]
    \centering
    \includegraphics[width=\textwidth]{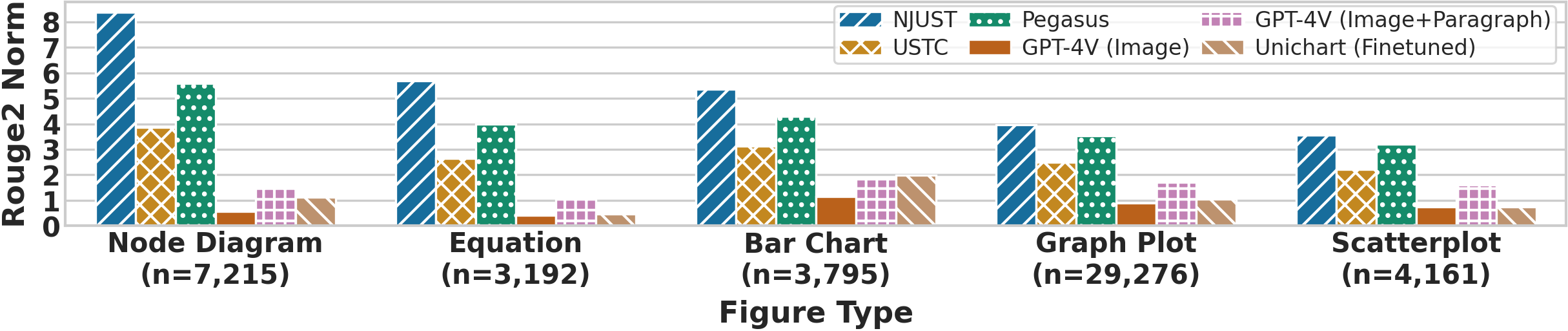}
    \caption{ROUGE-2 scores by model across five figure types, showing similar trends.}
    \label{fig:figtype-rouge2_scores}
\end{figure*}

We conducted a series of analyses on each model's performance, including comparisons with advanced LMMs that were unavailable during the challenge. 
Study 1 (Section~\ref{subsec:model-included}) and Study 2 (Section~\ref{subsec:auto-eval-results}) assessed performance using the challenge's hidden test set, with Study 1 focusing on automatic evaluation results and Study 2 on human evaluation results. 
To verify the generalizability of our findings, Study 3 (Section~\ref{subsec:unseen-arxiv-eval}) repeated these evaluations on newer arXiv papers.


\subsection{Models Included in the Evaluation\label{subsec:model-included}}

This analysis compared the models of the two winning teams with three other sets of models:

\paragraph{Text Summarization Model: Pegasus.}
Prior studies showed that generating captions for scientific figures in scholarly articles can be effectively tackled by treating it as a text summarization task, \ie, summarizing all paragraphs mentioning the target figure into its caption~\cite{huang2023summaries}.
We fine-tuned Pegasus~\cite{zhang2020pegasus}, a well-known text summarization model, with the SciCap Challenge dataset's training set in our performance comparison.


\paragraph{\LMM (LMM): GPT-4V.\footnote{\label{note:model-version}We used \textit{gpt-4-vision-preview} in the main study. However, due to its deprecation on June 17, 2024, we employed \textit{gpt-4-0125-preview} in the~\Cref{subsec:img+text-textonly} and~\Cref{subsec:unseen-arxiv-eval}.}}
We evaluated GPT-4V for figure captioning in two settings:
{\em (i)} using \textbf{figure-mentioning paragraphs and images} as input, and
{\em (ii)} using \textbf{only figure images}.
We used GPT-4V through OpenAI's API.
\Cref{app:prompt} describes the details.


\paragraph{Open LMM: UniChart~\cite{masry2023unichart}.}
We fine-tuned UniChart 
using the \dataset Challenge data. 
The training and inference details are provided in \Cref{app:model}.

\subsubsection{Why Exclude Text-Only LMMs?}
\label{subsec:img+text-textonly}
In our main study, we evaluated only LMMs that can take images as input, excluding text-only conditions. 
This decision followed our observation that combining text and images results in better captions. 
A human evaluation confirmed this performance difference:
We randomly sampled 200 figures from the arXiv Challenge Dataset (\Cref{sec:challenge}) and generated two captions for each figure using GPT-4V: one caption based solely on the textual context (\ie, paragraphs referencing the figure), and another that also incorporated the figure images. 
The same three experts who performed human evaluation in Study 2 (\Cref{subsec:human-eval-results}) ranked these captions. 
Experts ranked image+text captions higher than text-only captions in 82.5\%, 66\%, and 56\% of cases, respectively. 
This indicates that captions with images (mean=1.32; lower scores denote better performance) significantly outperformed text-only captions (mean=1.68).

\subsection{Study 1: Automatic Evaluation Results\label{subsec:auto-eval-results}}

We evaluated our models using BLEU-4, ROUGE-1, ROUGE-2, and ROUGE-L metrics~\cite{papineni2002bleu,lin-2004-rouge}, calculating ROUGE scores with the rouge-score tool on all-lowercase, stemmed text~\cite{rouge-scorer}.
Following \citet{huang2023summaries}, we also used normalized ROUGE scores because ROUGE can be affected by text length, with longer texts typically scoring higher~\cite{sun-etal-2019-compare}.
The normalization factor was computed on the hidden test set.
\Cref{tab:results_by_domain} and \Cref{tab:results_by_type} show model performance across domains and figure types.


\paragraph{Models maintained consistent performance rankings across categories.}
\Cref{fig:domains-rouge2_scores} and \Cref{fig:figtype-rouge2_scores} showed the ROUGE-2-Normalized scores for each model across various domains and figure types.
The Leaderboard Winner, NJUST, consistently scored the highest in every category, followed by PEGASUS in second place and USTC in third. 
Scores for both LLMs and LMMs were consistently lower. 
Model score rankings were consistent across categories, but score ranges varied. 
The domains of \texttt{cs} and \texttt{eess} saw the highest scores, and Node Diagrams emerged as the figure type with the highest scores.

We speculate that differences in automatic scores across categories stem from the linguistic characteristics of captions rather than the amount of training data available. 
\texttt{physics}, the domain with the lowest scores in Figure~\ref{fig:domains-rouge2_scores}, was the most common, and graph plots, the most frequent figure type, also scored low in Figure~\ref{fig:figtype-rouge2_scores}.

\subsection{Study 2: Human Evaluation Results By Professional Editors\label{subsec:human-eval-results}}
\begin{figure*}[t]
    \centering
    \includegraphics[width=0.98\linewidth]{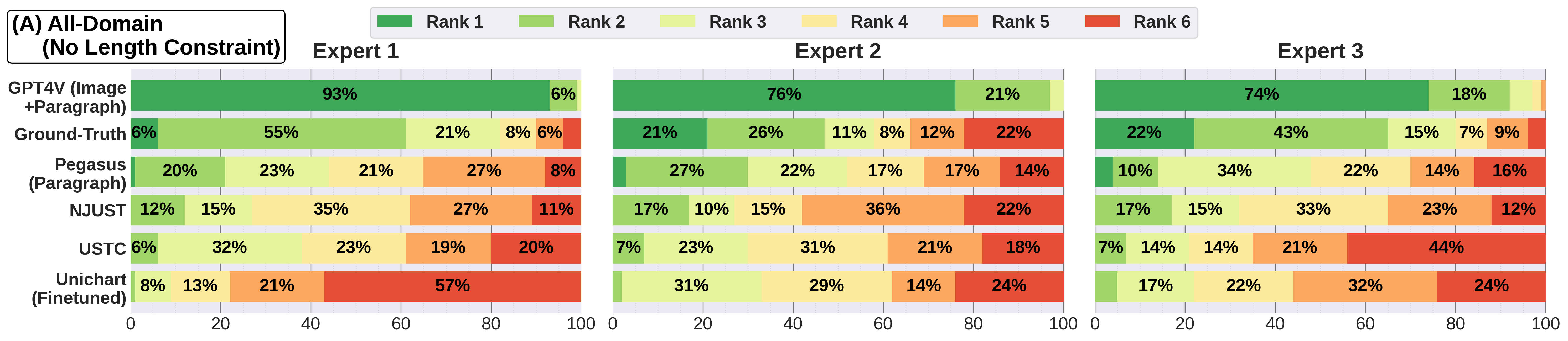}
    \includegraphics[width=0.98\linewidth]{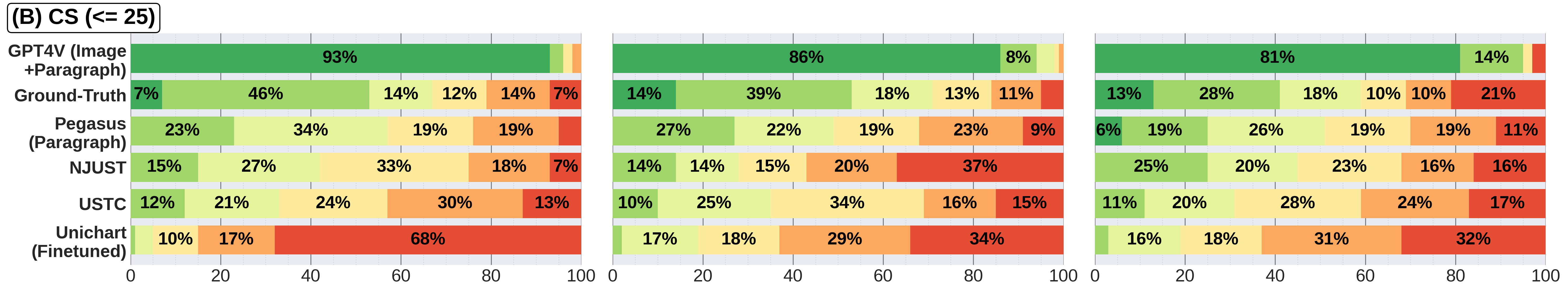}
    \includegraphics[width=0.98\linewidth]{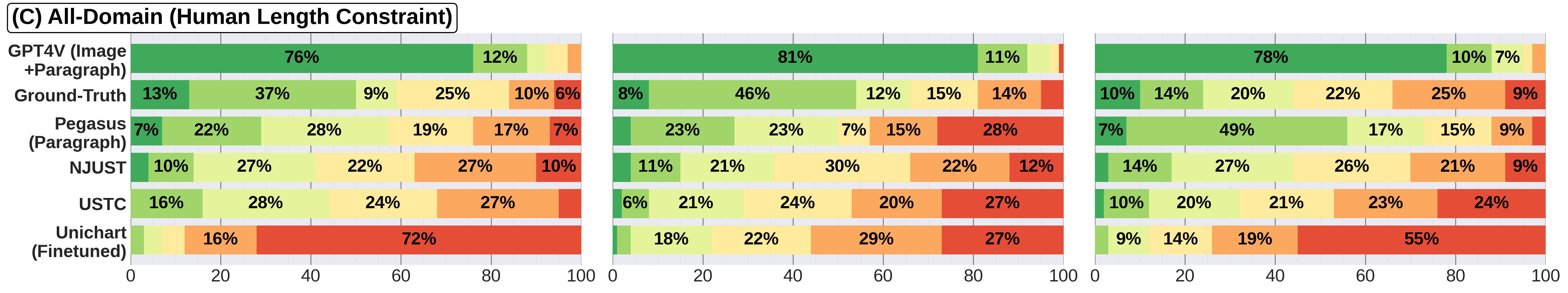}
\caption{Rankings of generated captions by all models in Study 2 across three evaluation conditions (A, B, C) and three experts. 
Models were ranked from 1 (best, green) to 6 (worst, red). 
GPT-4V (Image+Paragraph) consistently outperformed other models, including humans, across varying length constraints: none (A), a 25-word limit (B), and a strict limit matching human-written caption length (C).}
    \label{fig:main-human-eval-results}
\end{figure*}

\paragraph{Expert Judges.}
One of the primary goals of the \dataset dataset and challenge is to produce captions that help human readers.
To assess this, we recruited three professional editors through UpWork (upwork.com),
specializing in academic articles in the technical field and all native American English speakers. 
Their qualifications include: 
one with over ten years of editing experience and a Ph.D. in Comparative Literature, and two from STEM fields---one in Theoretical Astrophysics and another in Neuroscience---both with a significant track record in editing, proofreading, and publishing academic papers. 
They spent between 30 to 60 minutes evaluating 10 figures, with their rates ranging from \$50 to \$60 per hour. 

\paragraph{Three Varying Length Constraints.}
These editors conducted a three-part human evaluation. 
Caption length is known to be a key factor influencing human judgments of quality, as longer captions are often perceived as more informative by readers~\cite{hartley2003single,gelman2002let}.\footnote{Note that we did not deliberately control the generation lengths in Study 1, as the length biases in automatic metrics were taken care of by metrics normalization.}
To assess human-perceived model performance under varying length constraints, we conducted human evaluations for captions generated under three different settings, ranging from no length constraints to the strictest length constraints:


\begin{enumerate}


\item \textbf{Generation with no length constraints (all arXiv domains, Figure~\ref{fig:main-human-eval-results}A):}
For the first part, we selected 100 figures from the hidden test set and asked each editor to rank 6 captions for each figure:
1 author-written and 5 machine-generated by models used in Section~\ref{subsec:auto-eval-results} (including the two teams' outputs), excluding GPT-4V with image-only input due to its poor quality. 
They used a drag-and-drop interface (Figure~\ref{fig:human-ui} in Appendix~\ref{app:interface}), similar to that of~\citeauthor{hsu2023gpt}, to rank the captions based on the criterion: ``When I read the paper, the caption can help me understand the message that the figure tries to convey.''
This represented the least length-constrained setting.
For additional context, when no length constraints were specified in prompts, the average length of GPT-4V's captions (34.9 words) was shorter than that of human-written captions (42.9 words).

\item 
\textbf{Generation with 25-word length constraints (\texttt{cs} Papers, Figure~\ref{fig:main-human-eval-results}B):}
The second part followed the same procedure as the first, but focused specifically on 100 figures from \texttt{cs} domain papers. 
We selected the \texttt{cs} domain because the average length of human-written captions in \texttt{cs} arXiv papers is 25.52 words, shorter than GPT-4V's average caption length (33.6 words) when no length constraints are given. 
This choice helped mitigate the potential effect of longer human-written captions in the first condition. 
To align GPT-4V's captions with the average length in the \texttt{cs} domain, we modified the prompt to limit caption length to 25 words. 
Consequently, the average length of GPT-4V's captions was 25.79 words, closely matching the 25-word target.


\item 
\textbf{Generation with length no longer than human captions (all arXiv domains, Figure~\ref{fig:main-human-eval-results}C):}
In the third part, we followed the same procedure as in the first condition to sample 100 figures from papers of all arXiv domains. 
We modified GPT-4V's prompt to restrict caption length to be no longer than the corresponding human-written captions from arXiv papers, establishing the strictest length constraints among all three settings.
In this setting, generated captions were expected to always be shorter or match the length of human-written captions.



\end{enumerate}

\begin{figure*}[t]
    \centering
    \includegraphics[width=0.98\linewidth]{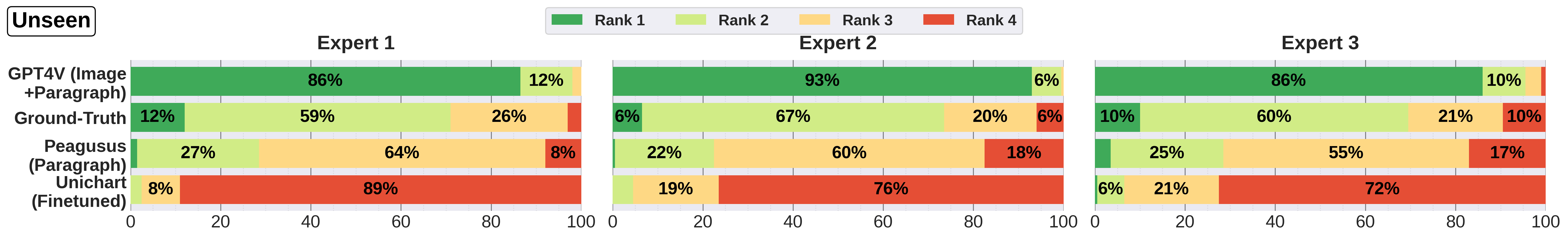}
\caption{Rankings of captions generated by different models on an unseen dataset of arXiv papers posted after GPT-4V's data cut-off date, evaluated by three experts. The procedure replicated Study 2.}
    \label{fig:unseen-human-eval-results}
\end{figure*}

\paragraph{GPT-4V captions were preferred over author-written ones and all other models.}
The ranking results are shown in \Cref{fig:main-human-eval-results}. 
Captions generated by GPT-4V, using both paragraphs and images, were overwhelmingly preferred by all experts in every setting, surpassing even author-written captions. 
Although author-written captions ranked slightly lower than GPT-4V, they still outperformed all other models, remaining reliable when considering the proportion of outputs ranked in the top two. 
Among the remaining models, the text-only summarization model, Pegasus, consistently outperformed Unichart and the two team models. 
Unichart, which was designed for chart understanding and reasoning rather than text generation, received the lowest scores in human evaluation. 
It is worth emphasizing that extensive evidence shows that automatic evaluation metrics, like BLEU, do not align well with human judgment~\cite{dhingra-etal-2019-handling}.
Therefore, in this paper, we place greater emphasis on human evaluation over automatic evaluation results.
We also analyzed the experts' free-text feedback to understand their rankings, as detailed in Section~\ref{sec:comment-analysis}.


\paragraph{Length constraints did not affect GPT-4V's superiority.}
The human evaluation results show that, although human readers tend to favor longer captions, imposing stricter length constraints did not diminish the strong preference for GPT-4's captions. 
Each expert displayed individual preferences,
but overall, they favored GPT-4's captions in most cases.
Notably, GPT-4 roughly adhered to length constraints: 
In Condition 1 (all arXiv domains), the average lengths were 41.69 for human-written captions and 33.6 for GPT-4. 
In Condition 2 (\texttt{cs} domain only), the averages were 25.52 for humans and 25.79 for GPT-4. 
In Condition 3 (all arXiv domains with human-written length constrained), the human average was 39.68, while GPT-4's average was 44.59.


\paragraph{Inter-Annotator Agreements.}
We measured inter-annotator agreements between experts using Kendall's Tau (for ranking all six items) and Cohen's Kappa (for selecting the top 1 or top 2 items). 
The Kendall's Tau values for Condition A, B, and C across expert pairs (1-2, 2-3, 3-1) were (0.71, 0.68, 0.64), (0.66, 0.53, 0.59), and (0.6, 0.58, 0.56), respectively, indicating a strong positive correlation between experts.
Given that our caption ranking study prioritizes the selection of top-ranked captions---since users are more likely to adopt these as drafts or read them---we also evaluated agreement using Cohen's Kappa, treating the ranking task as a binary classification task: captions ranked first were labeled as 1, and all others as 0 (Kappa@1). 
Similarly, Kappa@2 classified the top two captions.
For Kappa@1, the values for Conditions A, B, and C were (0.74, 0.64, 0.70), (0.78, 0.81, 0.72), and (0.63, 0.59, 0.53), respectively. 
For Kappa@2, the corresponding values were (0.72, 0.78, 0.72), (0.66, 0.60, 0.52), and (0.61, 0.55, 0.58). 
These results indicate substantial agreement between experts.

\subsection{Study 3: Are Results Generalizable to Papers Posted After the Cut-off Date?\label{subsec:unseen-arxiv-eval}}

\begin{table*}[ht!]
\centering
\mysmall
\begin{tabular}{@{}lcS[table-format = 2.1, detect-all]ccccccc@{}}
\toprule
\multirow{2}{*}{\textbf{Model}} & \multirow{2}{*}{\textbf{Domain}} & \multicolumn{1}{c}{\multirow{2}{*}{\shortstack{\textbf{Token}\\\textbf{Length}}}}
 & \multirow{2}{*}{\textbf{BLEU-4}} & \multicolumn{2}{c}{\textbf{ROUGE-1}} & \multicolumn{2}{c}{\textbf{ROUGE-2}} & \multicolumn{2}{c}{\textbf{ROUGE-L}} \\ \cmidrule(lr){5-6} \cmidrule(lr){7-8} \cmidrule(lr){9-10}  
 &  &  &  & \textbf{Score} & \textbf{Norm} & \textbf{Score} & \textbf{Norm} & \textbf{Score} & \textbf{Norm} \\ \midrule
\multirow{8}{*}{\textbf{Pegasus}} & \textbf{physics} & 39.1 & \textbf{.096} & \textbf{.378} & 1.651 & \textbf{.208} & 2.804 & \textbf{.311} & 1.864 \\
 & \textbf{cs} & 18.9 & \textbf{.054} & \textbf{.316} & 1.613 & \textbf{.157} & 2.674 & \textbf{.275} & 1.796 \\
 & \textbf{math} & 22.1 & \textbf{.055} & \textbf{.321} & 1.563 & \textbf{.168} & 2.684 & \textbf{.282} & 1.791 \\
 & \textbf{q-bio} & 29.6 & \textbf{.067} & {\ul .317} & 1.475 & \textbf{.162} & 2.415 & \textbf{.272} & 1.690 \\
 & \textbf{stat} & 25.0 & \textbf{.061} & {\ul .339} & 1.622 & \textbf{.178} & 2.757 & \textbf{.289} & 1.819 \\
 & \textbf{q-fin} & 21.4 & \textbf{.051} & \textbf{.327} & 1.606 & \textbf{.166} & 2.684 & \textbf{.285} & 1.815 \\
 & \textbf{eess} & 19.6 & \textbf{.072} & \textbf{.344} & 1.736 & \textbf{.182} & 3.056 & \textbf{.303} & 1.964 \\
 & \textbf{econ} & 22.3 & \textbf{.027} & \textbf{.274} & 1.331 & \textbf{.124} & 1.976 & \textbf{.236} & 1.499 \\ \hline
\multirow{8}{*}{\shortstack[l]{\textbf{GPT-4V}\\\textbf{(Image+Paragraph)}}} & \textbf{physics} & 77.8 & {\ul .032} & {\ul .375} & 1.461 & {\ul .137} & 1.483 & {\ul .267} & 1.518 \\
 & \textbf{cs} & 53.3 & {\ul .018} & {\ul .307} & 1.225 & {\ul .110} & 1.301 & {\ul .240} & 1.362 \\
 & \textbf{math} & 60.8 & {\ul .018} & {\ul .297} & 1.170 & {\ul .112} & 1.283 & {\ul .237} & 1.340 \\
 & \textbf{q-bio} & 59.4 & {\ul .019} & \textbf{.331} & 1.303 & {\ul .117} & 1.350 & {\ul .250} & 1.415 \\
 & \textbf{stat} & 61.3 & {\ul .025} & \textbf{.345} & 1.360 & {\ul .132} & 1.508 & {\ul .264} & 1.491 \\
 & \textbf{q-fin} & 52.5 & {\ul .021} & {\ul .318} & 1.276 & {\ul .121} & 1.443 & {\ul .251} & 1.430 \\
 & \textbf{eess} & 59.1 & {\ul .017} & {\ul .290} & 1.145 & {\ul .103} & 1.186 & {\ul .229} & 1.292 \\
 & \textbf{econ} & 57.9 & {\ul .016} & {\ul .257} & 1.012 & {\ul .099} & 1.140 & {\ul .211} & 1.190 \\ \hline
\multirow{8}{*}{\shortstack[l]{\textbf{Unichart}\\\textbf{(Finetuned)}}} & \textbf{physics} & 25.2 & .004 & .171 & 0.816 & .045 & 0.695 & .143 & 0.903 \\
 & \textbf{cs} & 9.1 & .004 & .176 & 1.289 & .059 & 1.590 & .159 & 1.354 \\
 & \textbf{math} & 15.5 & .002 & .152 & 0.834 & .040 & 0.746 & .136 & 0.930 \\
 & \textbf{q-bio} & 11.7 & .004 & .165 & 1.037 & .052 & 1.152 & .146 & 1.101 \\
 & \textbf{stat} & 13.7 & .003 & .175 & 1.018 & .056 & 1.120 & .154 & 1.100 \\
 & \textbf{q-fin} & 11.8 & .003 & .185 & 1.161 & .067 & 1.479 & .166 & 1.253 \\
 & \textbf{eess} & 11.9 & .007 & .201 & 1.250 & .076 & 1.663 & .183 & 1.373 \\
 & \textbf{econ} & 14.5 & .008 & .148 & 0.839 & .047 & 0.906 & .133 & 0.928 \\ \bottomrule
\end{tabular}
\caption{Results breakdown by domain for each model, highlighting performance metrics such as BLEU-
4, ROUGE scores, and token length. We show the models included in the human evaluation process. We test the finetuned model in~\Cref{tab:results_by_domain} on unseen data. The \textbf{highest} and \underline{second highest} values across models are highlighted.}
\label{tab:autometrics_results_by_domain_unseen}
\end{table*}

\begin{table*}[ht!]
\centering \mysmall
\begin{tabular}{@{}lcS[table-format = 2.1, detect-all]ccccccc@{}}
\toprule
\multirow{2}{*}{\textbf{Model}} & \multirow{2}{*}{\textbf{Domain}} & \multicolumn{1}{c}{\multirow{2}{*}{\shortstack{\textbf{Token}\\\textbf{Length}}}}
 & \multirow{2}{*}{\textbf{BLEU-4}} & \multicolumn{2}{c}{\textbf{ROUGE-1}} & \multicolumn{2}{c}{\textbf{ROUGE-2}} & \multicolumn{2}{c}{\textbf{ROUGE-L}} \\ \cmidrule(lr){5-6} \cmidrule(lr){7-8} \cmidrule(lr){9-10}  
 &  &  &  & \textbf{Score} & \textbf{Norm} & \textbf{Score} & \textbf{Norm} & \textbf{Score} & \textbf{Norm} \\ \midrule
\multirow{5}{*}{\textbf{Pegasus}} & \textbf{Node Diagram} & 18.3 & \textbf{.056} & \textbf{.303} & 1.564 & \textbf{.160} & 2.770 & \textbf{.270} & {\color[HTML]{1F1F1F} \textbf{1.780}} \\
 & \textbf{Equation} & 22.3 & \textbf{.050} & \textbf{.303} & 1.473 & \textbf{.146} & 2.335 & \textbf{.263} & 1.669 \\
 & \textbf{Graph Plot} & 30.2 & \textbf{.080} & \textbf{.357} & 1.655 & \textbf{.190} & 2.812 & \textbf{.301} & 1.867 \\
 & \textbf{Scatterplot} & 35.1 & \textbf{.080} & {\ul .359} & 1.609 & \textbf{.186} & 2.612 & \textbf{.294} & 1.790 \\
 & \textbf{Bar Chart} & 21.0 & \textbf{.056} & {\ul .330} & 1.632 & \textbf{.161} & 2.635 & \textbf{.284} & 1.815 \\ \hline
\multirow{5}{*}{\shortstack[l]{\textbf{GPT-4V}\\\textbf{(Image+Paragraph)}}} & \textbf{Node Diagram} & 56.8 & {\ul .012} & {\ul .279} & 1.102 & {\ul .093} & 1.083 & {\ul .218} & 1.228 \\
 & \textbf{Equation} & 49.7 & {\ul .014} & {\ul .279} & 1.138 & {\ul .092} & 1.123 & {\ul .218} & 1.256 \\
 & \textbf{Graph Plot} & 68.9 & {\ul .027} & {\ul .346} & 1.354 & {\ul .128} & 1.423 & {\ul .257} & 1.457 \\
 & \textbf{Scatterplot} & 70.8 & {\ul .028} & \textbf{.366} & 1.433 & {\ul .133} & 1.471 & {\ul .263} & 1.491 \\
 & \textbf{Bar Chart} & 51.9 & {\ul .023} & \textbf{.338} & 1.360 & {\ul .126} & 1.511 & {\ul .263} & 1.500 \\ \hline
\multirow{5}{*}{\shortstack[l]{\textbf{Unichart}\\\textbf{(Finetuned)}}} & \textbf{Node Diagram} & 6.8 & .001 & .120 & 1.085 & .031 & 1.088 & .111 & 1.128 \\
 & \textbf{Equation} & 9.9 & .000 & .109 & 0.751 & .023 & 0.572 & .101 & 0.819 \\
 & \textbf{Graph Plot} & 19.7 & .006 & .189 & 0.954 & .060 & 1.003 & .164 & 1.063 \\
 & \textbf{Scatterplot} & 22.4 & .002 & .171 & 0.829 & .044 & 0.699 & .145 & 0.920 \\
 & \textbf{Bar Chart} & 9.9 & .004 & .202 & 1.400 & .072 & 1.807 & .181 & 1.473 \\ \bottomrule
\end{tabular}
\caption{Results are broken down by domain for each model, highlighting performance metrics such as BLEU-4, ROUGE scores, and token length. We include the models used in the human evaluation process and test the fine-tuned model in~\Cref{tab:results_by_type} on unseen data. The \textbf{highest} and \underline{second highest} values across models are highlighted.}
\label{tab:autometrics_results_by_type_unseen}
\end{table*}

A common concern with large pre-trained models is data contamination~\cite{balloccu-etal-2024-leak}, meaning that test data may overlap with data in the pretraining set. 
To address this, we repeated the experiments on \textbf{entirely unseen data for GPT-4V}, creating a new dataset from papers published after the model's data cut-off date, January 2024.

\paragraph{Experiment Setups.}
After filtering out tables and other figure types, this dataset initially contained 46,543 figures from 57,678 valid papers published between January and March 2024. 
Of these, 37,146 figures had associated paragraphs. 
We further refined the dataset by removing duplicate figure indices within the same paper, resulting in 36,841 figures.
Finally, after excluding figures with invalid captions, the dataset contained 36,740 figures, which is similar in size to the challenge's hidden test set.
We used \textit{gpt-4-0125-preview} here to generate figure captions (See~\Cref{note:model-version}). 

Similar to the challenge task, we conducted both automatic evaluation (\Cref{tab:autometrics_results_by_domain_unseen} and \Cref{tab:autometrics_results_by_type_unseen}) and human evaluation (\Cref{fig:unseen-human-eval-results}).
Note that two teams' models were excluded, as this study was not part of the \dataset Challenge.

\paragraph{The results from Studies 1 and 2 generalize well to unseen arXiv figures.}
As discussed in Study 2, this work places a greater emphasis on human evaluation. 
The human evaluation results from Study 3 (\Cref{fig:unseen-human-eval-results}) showed a similar trend to those in Study 2 (\Cref{fig:main-human-eval-results}), with GPT-4V's captions consistently favored by all experts. 
Additionally, Study 3's automatic evaluation results (\Cref{tab:autometrics_results_by_domain_unseen} and \Cref{tab:autometrics_results_by_type_unseen}) displayed trends similar to those in Study 1 (\Cref{tab:results_by_domain} and \Cref{tab:results_by_type}), where text-summarization models received higher scores. 
Overall, the findings from Studies 1 and 2 appear to generalize to figures in unseen arXiv papers.

\section{Additional Study: Do Paper Readers Agree with the Editors' Judgement?\label{sec:student-ratings}}


Study 2 (Section~\ref{subsec:human-eval-results}) shows that professional editors' perspective toward machine-generated captions, 
raising an intriguing question: 
Do paper \textit{readers} evaluate caption quality in the same way as professional editors?
This distinction is important because prior studies have often relied on general readers, such as graduate students, rather than professional editors for human evaluations~\cite{petsiuk2022humanevaluationtexttoimagemodels,kasai-etal-2022-transparent}.
Understanding any systematic differences between readers and editors is essential to inform the broader research community.

\begin{figure}[t]
    \centering
    \includegraphics[width=0.98\linewidth]{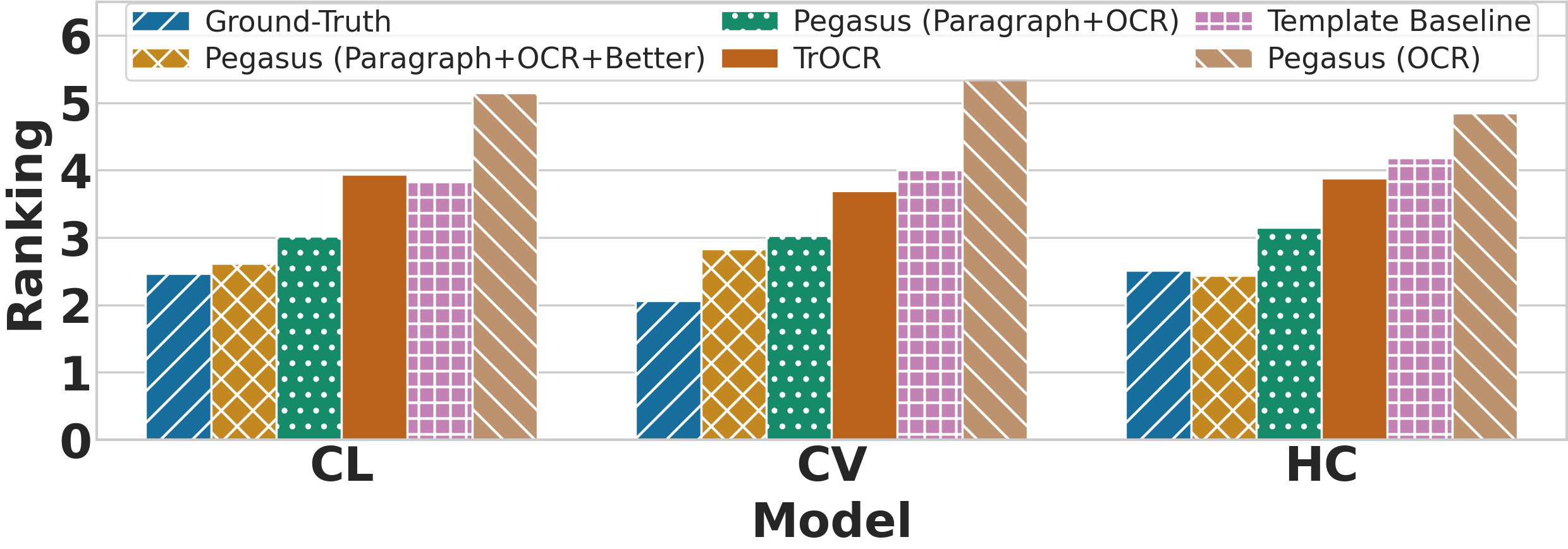}
    \caption{Ph.D. students' ranking results from~\citet{hsu2023gpt}. Note that lower ranks mean better performance.}
    \label{fig:category-comparison-expert}
\end{figure}

\begin{figure}[t]
    \centering
    \includegraphics[width=0.98\linewidth]{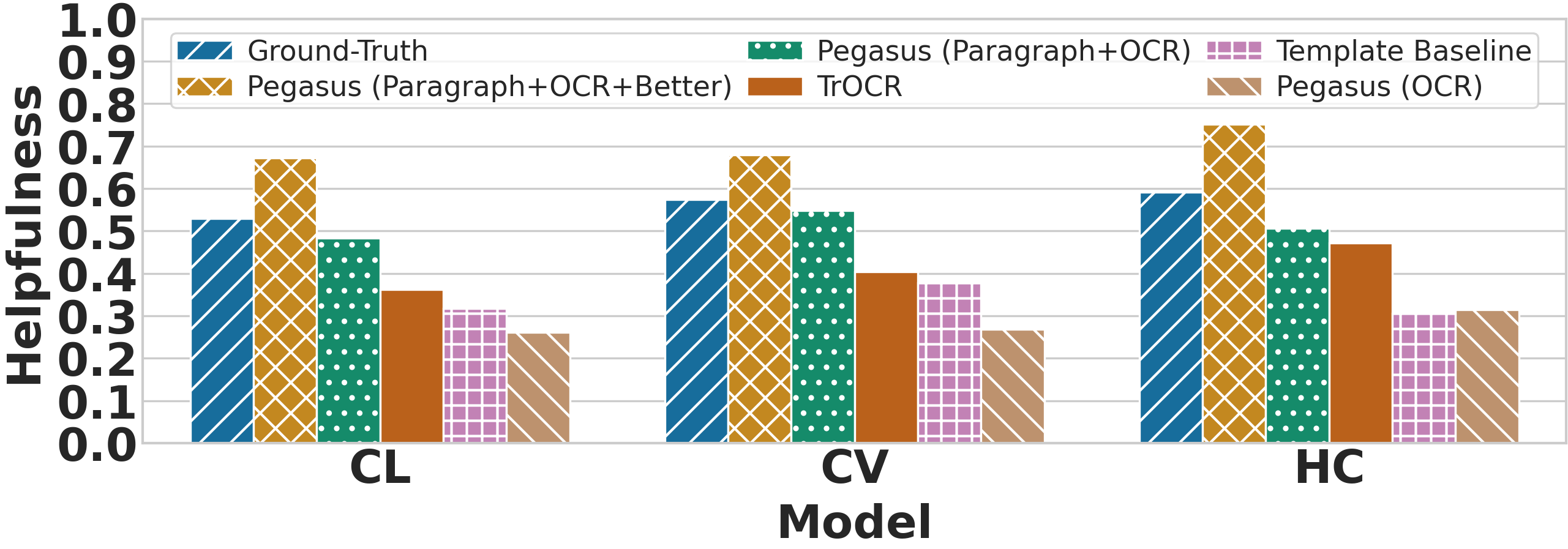}
    \caption{Undergraduate students' ratings on helpfulness from~\citet{hsu2023gpt}. Note that higher ratings mean better performance.}
    \label{fig:category-comparison-undergrad}
\end{figure}

To answer this question, we obtained the data for the study conducted by~\citet{hsu2023gpt}, where Ph.D. students in the relevant fields ranked figure captions generated by different models, and undergraduate students rated the captions' helpfulness.
The models \citet{hsu2023gpt} used 
differed from ours but similarly employed a fine-tuned text summarization model, Pegasus, as a baseline. 
Their study included three versions of captions produced by Pegasus: 
one fine-tuned with the entire training set of the \dataset dataset [Pegasus (Paragraph+OCR)], 
one fine-tuned using only captions exceeding 30 words in length [Pegasus (Paragraph+OCR+Better)],\footnote{It was called ``Better'' because previous studies indicated that longer captions enhanced reader comprehension~\cite{hartley2003single,gelman2002let}.}
and the other fine-tuned using only the figure image's OCR [Pegasus (OCR)].
They also included author-written captions in their comparison, providing a basis for comparing their results with ours.
Interestingly, their study focused on three \texttt{cs} subdomains of arXiv: 
natural language processing (NLP, \texttt{cs.CL}), 
human-computer interaction (HCI, \texttt{cs.HC}), and 
computer vision (CV, \texttt{cs.CV}).
This section describes the insight gained from comparing their results, as shown in Figure~\ref{fig:category-comparison-expert} and~\ref{fig:category-comparison-undergrad}, with ours.

\paragraph{Students agreed with editors that the \textit{basic} text-summarization model did not outperform paper authors.}
Ph.D. students (Figure~\ref{fig:category-comparison-expert}) and undergraduate students (Figure~\ref{fig:category-comparison-undergrad}) both preferred author-written captions to those generated by the Pegasus (Paragraph+OCR) model, which was fine-tuned on the entire dataset rather than just long captions.
These preferences aligned with the editors' judgments (Section~\ref{subsec:human-eval-results}), where human-authored captions were chosen more often over Pegasus-generated ones, even though our text-summarization model was fine-tuned on a much larger dataset than \citet{hsu2023gpt}'s.

\paragraph{User groups do not always agree with each other.}
While the overall trend in quality assessment among models was consistent between Ph.D. students and undergraduates---such as the high rating of human-written captions---agreement was not universal.
Undergraduates found captions from the Pegasus model trained on longer captions more useful than human-written ones (Figure~\ref{fig:category-comparison-undergrad}), a view not shared by Ph.D. students (Figure~\ref{fig:category-comparison-expert}). 
This discrepancy shows diverse user needs~\cite{gkatzia-etal-2014-finding}, beyond what editors' ratings capture.

\section{Discussion\label{sec:discussion}}

\begin{figure*}[t]
    \centering
    \includegraphics[width=0.99\linewidth]{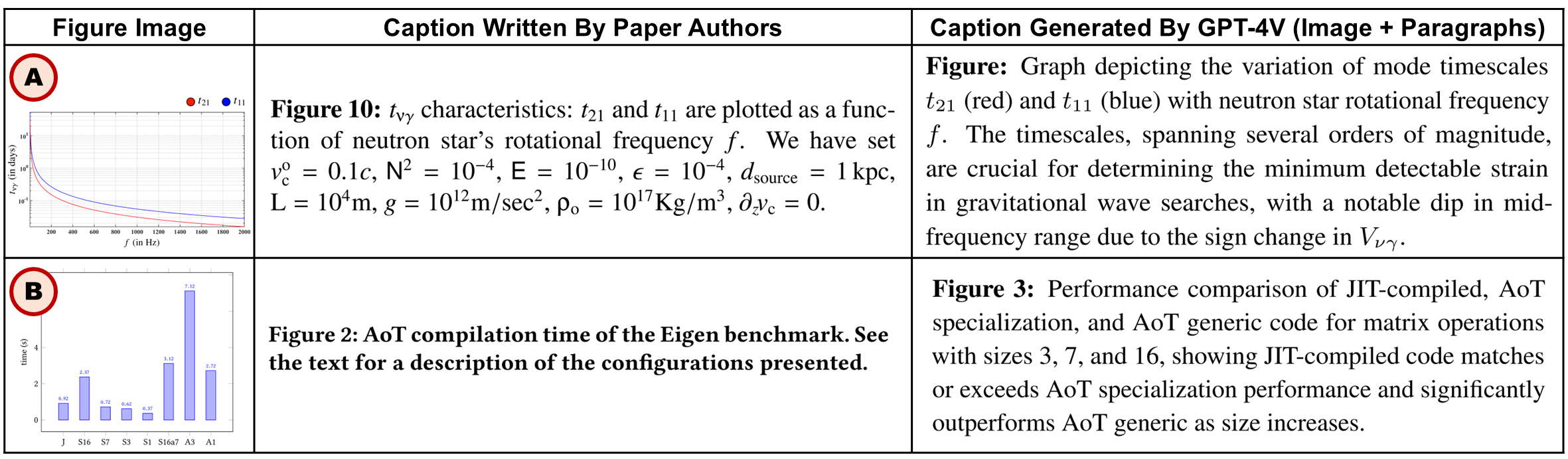}
    \caption{Two examples where experts favored GPT-4V's captions for providing sufficient details and highlighting key takeaway messages, two key factors identified in our comment analysis. [Figure sources: (A)~\citet{singh2017gravitational}, (B)~\citet{finkel2019clangjit}]}
    \label{fig:gpt-vs-humans}
\end{figure*}


\subsection{Is the Problem Solved?}
Our human evaluations show that GPT-4V often generates captions that are seen as better than those written by authors (Section~\ref{subsec:human-eval-results}). 
However, this does not fully solve the challenge of generating captions for scientific figures. 
We have reached an important milestone, but there are new challenges ahead:

\begin{itemize}
    \item 
First, \textbf{improving caption quality further}. 
Being better than human-written captions does not necessarily mean the captions are of high quality; it might simply mean that the human-written captions are bad. 

\item
Second, \textbf{evaluation is still hard.}
Despite GPT-4V's poor automatic scores (Section~\ref{subsec:auto-eval-results}), it outperformed other models in evaluations by editors;
efforts to use a higher-quality subset for better evaluation did not significantly differ in scores from the full test set (Table~\ref{leaderboard}). 
These highlight the need for more research into evaluation methods, particularly as the quality of machine-generated content surpasses that of human-generated content. 
Future research should focus on detail, coherence, and factual accuracy. 
Developing a reliable automatic evaluation method could also enhance the quality of datasets used to develop new models.

\item 
Finally, \textbf{personalization for different users}.
Different user groups have different information needs. 
Another challenge is to create models that can customize captions to meet the diverse preferences of various users.

\end{itemize}

\subsection{Generalizability of Caption Generation Models}
Our findings suggest caption generation models are versatile across domains and figure types.
Although there are some variations and domain-specific details, \lmms pre-trained on extensive data can likely generate usable captions for many domains and figure types. 
We also tested several hypotheses to explain the variations in automatic scores across domains, such as the presence of non-English characters in captions or length of captions, but found no conclusive evidence to support them.

\subsection{What Makes GPT-4's Captions Better?\label{sec:comment-analysis}}
To understand why GPT-4's generated captions were often rated higher than author-written captions, we manually analyzed the comments provided by three experts during the ranking process. 
Although experts were instructed to provide comments, we did not strictly enforce this for every figure.
In total, we collected 1,123 comments corresponding to 500 figures.
The first author of this paper manually coded these comments using open coding practices to identify the reasons behind GPT-4V's superior ratings. 
Two key themes emerged: 
{\em (i)} providing \textbf{sufficient details} and 
{\em (ii)} highlighting the figure's \textbf{takeaway messages}. 
Among the 1,123 comments analyzed, 277 (24.7\%) indicated that captions were rated higher or lower based on the presence or absence of rich details, while 177 (15.8\%) noted that including or omitting takeaway messages influenced their ratings.
These findings align with prior research showing that detailed captions with clear takeaway messages receive higher ratings~\cite{hsu2023gpt}.
Because humans tend to prefer longer captions, we further analyzed the 207 comments from Study 2's C condition, where GPT-4's captions were restricted to the same length as the authors'.
Even under this constraint, the two key factors remained prominent: 
49 (23.7\%) comments mentioned details, and 50 (24.2\%) comments cited takeaway messages.
Figure~\ref{fig:gpt-vs-humans} shows two representative examples where experts stated that GPT-4V's captions were better due to these factors.

We also analyzed the experts' comments to identify the most common errors in AI-generated captions, particularly those from models other than GPT-4. 
These errors fell into three main categories: 
The first category is \textbf{incorrect visual information}, where visual details in captions, such as color-number relationships, are inaccurately described. 
The second category involves \textbf{linguistic errors}, including typos, missing punctuation, or minor grammatical mistakes. 
The final category is \textbf{factual errors unrelated to visuals}, such as incorrect numerical values.
In addition to outright errors, we also examined cases where captions, though not strictly incorrect, were problematic. 
One common issue was \textbf{incomplete captions}, where truncation resulted in phrases like ``Delivery ratio vs.'' or ``Plot of F2 versus ab.'' 
Another problem was \textbf{repetitive or overly generic captions}, where models produced identical or indistinct descriptions, such as ``IoT Hub Workflow,'' reflecting a lack of specificity.

\section{Conclusion and Future Work\label{sec:conclusion}}
This paper presents the first \dataset Challenge results, highlighting GPT-4V's superiority in generating scientific figure captions, as professional editors overwhelmingly preferred figure captions generated by GPT-4V over those from all other models and even the original captions written by authors.
Despite this achievement, we acknowledge that while an important milestone has been reached in caption generation, challenges such as improving quality, evaluation, and customization remain unresolved.

\section{Limitations}
This work has several limitations.
First, our human evaluation involved only a small number of experts---three professional editors.
While this scale is not uncommon in NLP research due to the high cost of expert evaluation, it may not sufficiently capture the diverse perspectives and interpretations of a broader group of users or experts. 
Second, the comparison made in Section~\ref{sec:student-ratings} between our results and those from \citet{hsu2023gpt} was not fully direct due to differences in data and models, which we made clear in the paper.
While we believe the comparison is valid at a high level, it may not be broadly generalizable.
Third, our analysis spans all domains present on arXiv, but arXiv does not encompass all potential academic domains, such as biology and medicine, which are primarily published on PubMed. 
The characteristics of figures and captions in these areas may differ significantly from those in arXiv, potentially limiting the applicability of our conclusions outside of arXiv without additional verification.
Finally, our best-performing model, GPT-4V, is proprietary, with no public access to its architecture or training data. 
This restriction, common in NLP research, means our findings may inherit limitations related to the closed nature of the model, including potential data contamination issues, and lacks transparency for in-depth analysis.

\section{Ethics Statement}
We recognize that employing LLMs or LMMs to produce texts for end users inherently carries risks, including the dissemination of inaccurate or misleading information. 
In scholarly contexts, particularly when using generated captions to enhance figure comprehension, such inaccuracies could mislead readers. 
We have ensured to inform participants of these risks during our user studies.

\section*{Acknowledgments}
We sincerely appreciate the anonymous reviewers and action editors of Transactions of the Association for Computational Linguistics (TACL) for their valuable feedback. 
We also thank the human experts for their support in our evaluation efforts. 
Additionally, we express our gratitude to all the teams that participated in \textsc{SciCap} Challenge 2023 and to the organizers of the 5th Workshop on Closing the Loop Between Vision and Language (CLVL) at ICCV 2023 for hosting the \textsc{SciCap} Challenge 2023.
This research was partially supported by the Alfred P. Sloan Foundation (Grant Number: 2024-22721).

\bibliography{bibtex/tacl2021,bibtex/main,bibtex/anthology,bibtex/custom}
\bibliographystyle{acl_natbib}

\appendix

\section{\dataset Challenge Data Preparation}
\label{app:data-prepare}

The \dataset Challenge dataset was an expansion of the original \dataset dataset, incorporating papers from all 8 primary arXiv domains published between 2010 and 2020, totaling over 440,000 papers. 
Using PDFFigures 2.0~\cite{clark2016pdffigures}, 2.62 million figure-caption pairs were extracted from these papers.
The dataset also included paragraphs referencing these figures, extracted using GROBID,\footnote{GROBID: \url{https://github.com/kermitt2/grobid}} and identified figure-mentioning sentences (\eg, ``Figure 4 shows...'') through regular expressions.
The dataset featured OCR text from images obtained via Tesseract-OCR,\footnote{Tesseract-OCR: \url{https://github.com/tesseract-ocr/tesseract}} enhancing the data with legends and labels directly from figures.
A figure-separator tool~\cite{figure-separate} was used to spot compound figures;
FigureSeer~\cite{siegel2016figureseer} was used to classify figure types, excluding those categorized as ``Table'' and ``Others''. 
Finally, to accommodate teams with limited resources and ensure a balanced dataset, we limited the selection to four figures per paper, with figures from the same paper allocated to the same dataset split.

\paragraph{Additional Figures from ACL Papers.}
Additionally, the challenge dataset included 26,868 figures collected from ACL Anthology papers~\cite{rohatgi-etal-2023-acl} through ACL-Fig~\cite{karishma2023acl} as additional training data. 
In the ACL collection, we removed all the papers that were already published at arXiv. 
For this filtering, we relied on the paper clustering done by Semantic Scholar Academic Graph~\cite{Kinney2023TheSS}.

\section{Prompts Used}
\label{app:prompt}
In this section, we provide the prompt we used in \Cref{subsec:model-included}.
\texttt{[PARAGRAPHS]} and \texttt{[IMG]} are placeholders for figure-mentioning paragraphs and the encoded images.

\paragraph{Prompt without paragraph.}
\textit{``[IMG] Please write a short and concise caption for the figure.''}

\paragraph{Prompt with paragraph.}
\textit{``[IMG] Paragraphs: [PARAGRAPHS]. Above are a figure and referred paragraphs about the figure. Please write a short and concise caption for the figure.''}


\section{Model Details}
\label{app:model}
We describe the model training details and the decoding configuration used in Section \ref{subsec:model-included}.

\paragraph{Training Details for Open-Source LMMs.}
We fine-tune Unichart\footnote{We used \texttt{ahmed-masry/unichart-base-960}.} checkpoint from HuggingFace for 40000 steps, using batch size = 16, learning rate = 5e-5 with a linear decay scheduler, warmup steps = 500, weight decay = 0.01. We evaluate every 1000 steps, and the checkpoint with the highest highest ROUGE-2 score on validation set is kept and used to predict final result.

\paragraph{Decoding Details for Open-Source LMMs.}
For generation, captions were decoded using the beam sampling strategy, with beam size = 4, top-k = 50, and maximum length = 100.

\section{Interface for Human Evaluation\label{app:interface}}
\begin{figure*}[t]
    \centering
    \includegraphics[width=0.99\linewidth]{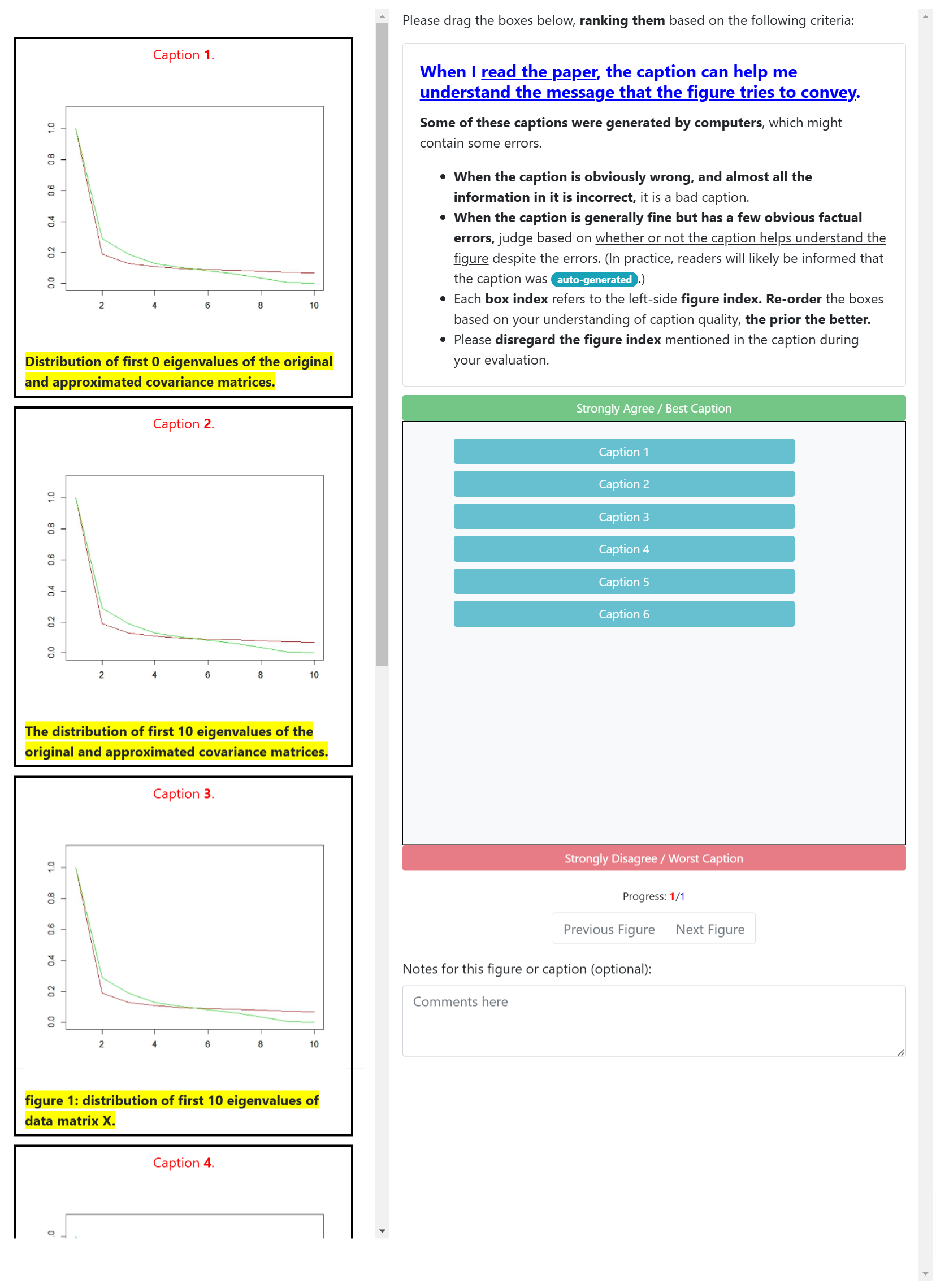}
    \caption{The drag-and-drop interface used by professional editors to rank captions for a figure. [Figure source:~\citet{bulgakov2016iterative}]}
    \label{fig:human-ui}
\end{figure*}

\Cref{fig:human-ui} shows the interface used for human evaluation.

\end{document}